\documentclass[sn-nature, iicol]{sn-jnl}

\usepackage{graphicx}%
\usepackage{multirow}%
\usepackage{amsmath,amssymb,amsfonts}%
\usepackage{amsthm}%
\usepackage{mathrsfs}%
\usepackage[title]{appendix}%
\usepackage{xcolor}%
\usepackage{textcomp}%
\usepackage{manyfoot}%
\usepackage{booktabs}%
\usepackage{algorithm}%
\usepackage{algorithmicx}%
\usepackage{algpseudocode}%
\usepackage{listings}%
\usepackage{epstopdf}

\usepackage{bm}
\usepackage{multirow}
\usepackage{booktabs}  
\usepackage{colortbl}  
\usepackage{natbib}
\usepackage{geometry}
\theoremstyle{thmstyleone}%
\theoremstyle{thmstyletwo}%

\theoremstyle{thmstylethree}%

\begin{document}

\title[Article Title]{Hierarchical learning control for autonomous robots inspired by central nervous system}

\author[1]{\fnm{Pei} \sur{Zhang}}\email{pei.zhang088@outlook.com}

\author[2]{\fnm{Zhaobo} \sur{Hua}}\email{Bobby\_1752635630@outlook.com}
\equalcont{These authors contributed equally to this work.}

\author*[3]{\fnm{Jinliang} \sur{Ding}}\email{jlding@mail.neu.edu.cn}
\equalcont{These authors contributed equally to this work.}

\affil*[1]{\orgdiv{State Key Laboratory of Synthetical Automation for Process Industries}, \orgname{Northeastern University}, \orgaddress{\street{WenHua Road}, \city{Shenyang}, \postcode{110819}, \state{Liaoning}, \country{China}}}
%


\abstract{Mammals can generate autonomous behaviors in various complex environments through the coordination and interaction of activities at different levels of their central nervous system. In this paper, we propose a novel hierarchical learning control framework by mimicking the hierarchical structure of the central nervous system along with their coordination and interaction behaviors. The framework combines the active and passive control systems to improve both the flexibility and reliability of the control system as well as to achieve more diverse autonomous behaviors of robots. Specifically, the framework has a backbone of independent neural network controllers at different levels and takes a three-level dual descending pathway structure, inspired from the functionality of the cerebral cortex, cerebellum, and spinal cord. We comprehensively validated the proposed approach through the simulation as well as the experiment of a hexapod robot in various complex environments, including obstacle crossing and rapid recovery after partial damage. This study reveals the principle that governs the autonomous behavior in the central nervous system and demonstrates the effectiveness of the hierarchical control approach with the salient features of the hierarchical learning control architecture and combination of active and passive control systems.}

\maketitle

\begin{figure*}[t]%
	\centering
	\includegraphics[width=0.9\textwidth]{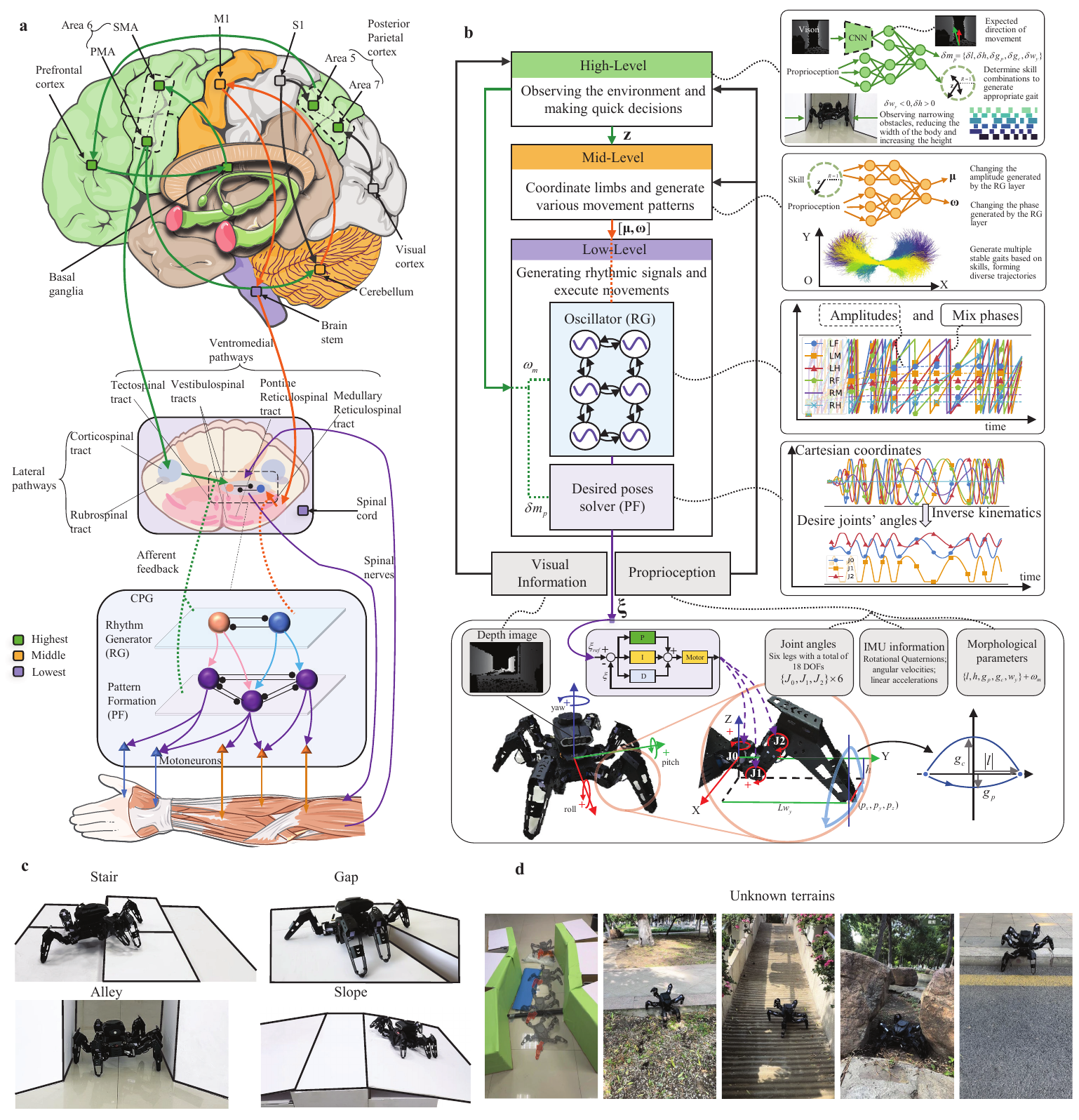}
	\caption{\textbf{Overview of the central nervous system and hierarchical learning control framework.} 
		\textbf{a}, Mammalian central nervous system structure, the figure contains the cerebral cortex partition, the spinal cord's internal structure, and the double-layer structure of CPGs neural circuits. 
		\textbf{b}, Schematic diagram of the proposed hierarchical control framework.
		The gray nodes in \textbf{a} and the gray box in \textbf{b} represent the sensing mechanism in the nervous system and control framework, respectively, and are responsible for the acquisition of sensing signals. In the nervous system, S1 and the visual cortex are mainly responsible. In the control frame, it is provided by sensor measurement. The green nodes and boxes in \textbf{a} and \textbf{b} represent the high-level institutions in the nervous system and control framework, respectively, responsible for observing the environment and making decisions. In the nervous system, most cortical regions are responsible for this function. In the control framework, this part is realized by the deep reinforcement learning neural network policy. The yellow nodes and boxes in \textbf{a} and \textbf{b} represent the mid-level institutions responsible for coordinating the limbs and generating various motion patterns. In the nervous system, the cerebellum and primary motor cortex are responsible. In the control framework, this part adopts an unsupervised reinforcement learning algorithm and skill-driven neural network. The purple nodes and boxes in \textbf{a} and \textbf{b} represent the low-level institutions that are responsible for the generation and execution of motion signals. In the nervous system, the brain stem and spinal cord are responsible. In the control framework, it is realized by the CPG module, which contains an oscillator and a desired pose solver to provide the desired joint position and uses the built-in PID feedback loop of the robot to control 18 motors. The solid line in \textbf{a} connects different nerve regions, representing the information flow relationship, and the thin purple solid line on the right represents the ascending and descending spinal nerves. Dotted lines indicate descending pathway feedback of the CPGs. The solid line in \textbf{b} represents the action relationship between the sensor and the control signal, and the black dotted line connects the specific analysis of each module. 
		\textbf{c}, Four different indoor obstacle terrain crossing tasks. 
		\textbf{d}, Various new obstacle terrain crossing tasks have never been learned.
	}\label{fig1}
\end{figure*}

Animals can develop a large number of motor skills to complete various complex tasks, which stem from the hierarchical structure of their central nervous system (CNS) and the complex interaction mechanisms between multiple regions\textsuperscript{\cite{bib1,bib2}}. Studying the hierarchical structure and mechanisms of the central nervous system can help understand the principle of autonomous movement and promote the design of autonomous robot control systems\textsuperscript{\cite{bib3,bib4}}.

Recently, approaches to developing robot autonomous control systems inspired by the central nervous system can be divided into two groups: hierarchical model active control approaches based on hierarchical structure and bionic passive control approaches based on neural mechanisms. The hierarchical model active control approaches aim to obtain a multi-level active control system with high flexibility and adaptability (consciously participated) by mimicking the hierarchical structure of the nervous system. Its internal controller continuously and actively provides control signals based on real-time feedback information to achieve specific goals \textsuperscript{\cite{bib5,bib6,bib7,bib8,bib9,bib10,bib11}}. These approaches generally include two stages: In the pre-training stage, the low-level controller learns low-level skills. In the task-training phase, the high-level controller learns to use these skills for autonomous decision-making\textsuperscript{\cite{bib5,bib6,bib7}}. However, finding effective reusable skills in complex robot systems is difficult. Learning low-level skills from data is a promising method, but the flexibility and versatility of skills are limited by the quality of data sets\textsuperscript{\cite{bib8,bib9}}. Collecting high-quality motion data for some special robots (such as insect robots) is also particularly difficult. The unsupervised reinforcement learning methods encourage robots to learn diverse skills through intrinsic objectives and eliminate the dependence on data. However, in complex systems with high degrees of freedom, it is usually unable to find useful skills\textsuperscript{\cite{bib10}}. The goal-driven methods transform low-level skill learning in complex systems into target tracking tasks by designing the target space\textsuperscript{\cite{bib11}}. Still, the manually designed target space (such as plane coordinates and joint angles) will limit the full exploration in the learning process. In addition, due to the need for real-time data feedback for continuous decision-making, active control systems rely heavily on sensor information.

The bionic passive control approaches aim to obtain a structurally simple and highly reliable passive control system (unconsciously participated) by simulating various mechanisms within the central nervous system. Through preset intrinsic internal mechanisms, the robot can spontaneously generate various behaviors, thereby overcoming many problems related to hierarchical model approaches. 
For example, by simulating the conditional reflex neural circuits inside the spinal cord (SC) of vertebrates and the ventral nerve cord (VNC) of invertebrates, control methods based on reflex mechanisms can directly and quickly trigger a series of rhythmic reflex behaviors\textsuperscript{\cite{bib12,bib13}} (such as crawling, etc.) based on sensory information, enabling robots to autonomously adapt to unpredictable irregular terrain\textsuperscript{\cite{bib14,bib15}}.
Central pattern generators (CPGs) are another type of neurons or neural circuits that exist within SC or VNC. They can generate periodic motor signals through internal oscillations without a bursting input and can be regulated by a small number of descending neurons in the brain\textsuperscript{\cite{bib16,bib17}}. 
By simulating this oscillation mechanism, CPG-based control methods have less dependence on sensory information and can enable robots to spontaneously generate rhythmic signals and form various biomimetic motion behaviors\textsuperscript{\cite{bib18,bib19}}  (such as swimming, walking, etc.). These methods rely on the inherent internal mechanisms of the system to generate motion behavior in robots, with simple structures and no need for complex real-time calculations. However, the simple structure also limits the flexibility of passive control systems.

Some studies use learning based methods combined with partial sensory feedback information to adjust the internal parameters of CPG controllers to achieve some simple target tasks\textsuperscript{\cite{bib20,bib21,bib22}}(such as target location navigation, speed tracking, etc.)
Although this semi-active control system enhances the flexibility of passive control, the control mode that only integrates CPG and simple descending drives is not conducive to learning complex motion skills and limits the diversity of autonomous behavior.

Among vertebrates, mammals have the most complete hierarchical central nervous system and exhibit highly autonomous, flexible, and reliable motor behaviors.
The developed spinal cord and complex reflex circuits in the central nervous system form a passive control system that can generate motion signals, while the highly evolved cerebral cortex forms an active control system capable of processing complex sensory information and making advanced decisions\textsuperscript{\cite{bib1}}. 
Analyzing the hierarchical structure of the central nervous system and the principles of sensory feedback, CPGs, and various descending drives mechanisms, and combining the advantages of active and passive control systems to design an autonomous robot control system, is expected to solve the problems of existing methods. Therefore, based on the analysis of the hierarchical collaboration relationship and functional mechanism of the mammalian central nervous system (Fig. \ref{fig1}a), we propose a hierarchical learning control framework for autonomous robots inspired by the central nervous system (Fig. \ref{fig1}b), which enables robots to acquire diverse and reusable skills, autonomously solve multiple tasks and adapt to unknown environments, and maintain their mobility in the absence of some sensing information. This framework combines the advantages of active and passive control systems, with flexibility and reliability.

Specifically, we make the following contributions to the design of autonomous robot control systems: 1) We analyze the working mechanism of the mammalian central nervous system, abstracts it as a three-layer semi-active control system, and describes the division of labor and cooperation relationships at different levels, providing a new perspective for the design of control systems. 2) We design and propose a semi-active hierarchical learning control framework based on the analysis results. The framework uses a learning based high-level neural network controller to mimic the cerebral cortex and basal ganglia, responsible for actively making decisions through sensory feedback and visual information; The framework uses a mid-level neural network controller to mimic the cerebellum and primary motor cortex, responsible for learning diverse and reusable skills and coordinating movements; The framework uses a low-level CPG module to mimic the brainstem and spinal cord, responsible for passively generating motion signals and being regulated by higher level controllers. By mimicking the descending drives mechanism of the spinal cord's lateral and ventromedial pathways, the high-level controller can actively control the lower level controllers through two control loops, thereby directly or indirectly adjusting the movement pattern. 3) We propose a CPG module embedded with independent phases, which enhances the system's fault tolerance and reduces the dependence of active control systems on sensing information. 4) We design a generalized skill pre-training method that references the learning mode of cerebellar motor skills, enabling the mid-level neural network controller to effectively explore reusable skills in high degree of freedom complex systems without the need for external data. 5) We design a multi-task learning method that references the planning patterns of movement in the cortex and basal ganglia, enabling the high-level neural network controller to quickly make autonomous decisions in various downstream tasks and unknown environments by using learned skills.

Following the proposed framework, we obtain a complete hierarchical controller and apply it to a complex hexapod robot PHAGE\textsuperscript{\cite{bibxiaoR}}. We observe that the PHAGE robot exhibit strong autonomous decision-making and flexible movement abilities in various obstacle crossing tasks, as well as rapid adaptive capabilities in unknown situations. Specifically, we designed various obstacle crossing tasks as shown in Fig. \ref{fig1}c (climbing stairs, crossing ravines, crossing narrow alleys, climbing slopes) to demonstrate animal level autonomous decision-making and movement abilities. In addition, we demonstrate the robot's ability to adapt to new environments and its rapid recovery capability under partial body damage conditions in various unknown obstacle environments shown in Fig. \ref{fig1}d. We also conduct comprehensive ablation studies in both simulation and real-world environments to evaluate the effectiveness of different components and demonstrate the advantages of the proposed framework.

The organizational structure of this paper is as follows: In the Results section, we show the working principle of the central nervous system and the detailed composition and performance evaluation results of the proposed framework. In the Discussion section, we discuss the similarity between the proposed control framework and the neural system and provide directions for further improvement and prospects for future work. In the Methods section, we provide specific implementation details.

\section{Results}\label{sec1}
We first present the analysis results of the mammalian central nervous system and provide an overview of our proposed framework based on the analysis results. Subsequently, in a bottom-up order, we evaluate the motion generation results of the CPG module within the framework, the skill learning and regulation results of the mid-level controller, and the multi-task learning and decision-making results of the high-level controller to demonstrate the effectiveness of different core components of the framework and the excellent autonomous decision-making and motion control capabilities of the proposed control method. Finally, through the rapid motion recovery effect of the robot under limb amputation conditions, as well as the rapid adaptation results of the robot in various unknown environments, we demonstrate that the proposed control framework has less dependence on sensor information and strong adaptive ability.

\subsection{Central nervous system inspired hierarchical learning control framework}\label{subsec11}

Combined with related research, we analyze the mammalian central nervous system. As shown in Fig. \ref{fig1}a, the central nervous system of mammals (taking humans as an example) is a three-level control system from the cerebral cortex to the spinal cord\textsuperscript{\cite{bib1,bib2}}. The highest level includes most cortical regions, such as the prefrontal cortex, premotor area (PMA), supplementary motor area (SMA), posterior parietal cortex, and basal ganglia. They are responsible for processing various information and generating high-level decisions. First, the ascending nerves of the spinal cord will project proprioception of the body to the primary somatosensory cortex (S1), and the visual nerves will project visual information of the retina to the visual cortex. Areas 5 and 7 of the posterior parietal cortex process proprioception from S1 and visual information from the visual cortex, respectively, and generate abstract perceptual information. Subsequently, the prefrontal cortex decides what action to take according to the perceptual information, and then the basal ganglia enhances these motor intentions through the internal direct pathway while inhibiting inappropriate motor programs through the indirect pathway\textsuperscript{\cite{bib26,bib27}}. This information mainly converges to the SMA of the cortex through the thalamus. Then, the SMA and PMA jointly carry out motor planning and control distal and postural muscles. The control information travels down the spinal cord through two main pathways\textsuperscript{\cite{bib2,bib24,bib25}}. On the one hand, through the ventromedial pathway, the high-level information is processed by the mid-level CNS composed of the cerebellum and primary motor cortex (M1) and transmitted to the brain stem and spinal cord at the lowest level to control the coordinated movement of limbs. In the above process, proprioception and decision-making information are projected from the pontine nucleus to the lateral cerebellum and then back to M1 through the thalamic ventral lateral nucleus (VLC) to generate finer motor signals and transmit them to the brainstem. Then, the postural muscles are controlled by spinal tracts such as the reticulospinal tract of the ventromedial column of the spinal cord. On the other hand, signals from SMA and PMA directly control the distal muscles through the corticospinal tract of the lateral column of the spinal cord. These afferent feedbacks will regulate CPGs neural circuits in the spinal cord, generating complex rhythmic signals to control the formation of autonomous movement of muscles. CPGs in the spinal cord have two levels of organization, in which the half-center rhythm generator layer (RG layer) is responsible for generating motor rhythms. The pattern formation layer (PF layer) is responsible for determining the exact form of muscle activation signals, mobilizing motoneurons to control muscles through spinal nerves\textsuperscript{\cite{bib28}}.

According to the above hierarchical structure and functional division of the central nervous system, we design the hierarchical learning control framework of Fig. \ref{fig1}b. The framework contains three levels of controllers from high to low. The high-level controller, which simulates the cortex and basal ganglia operation principle , is responsible for processing the proprioception and visual information provided by the sensors and making active and rapid decisions. Its internal convolutional neural network and somatosensory neural network process the depth image and proprioception, respectively, to analyze the environmental features, and speculate the next movement direction. The subsequent neural network processes such perceptual signals and provides high-level decisions. These decision vectors are divided into skill vectors and morphological parameters. The length of skill vectors determines the frequency of movement and simulates the regulation of basal ganglia on voluntary movement. They descend to the bottom CPG module through two control loops. The skill vectors are sent to the mid-level controller through the ventromedial loop. Inside the mid-level controller, the internal neural network combines the skills and proprioception to calculate the differential control signals. These signals are then sent to the CPG module, which generates various coordinated motion patterns. The morphological parameters are directly transmitted to the CPG module through the lateral loop, thus directly changing the shape of the limb (such as width and height). The CPG module comprises an oscillator and a desired pose solver, which simulate the functions of the RG and PF layers. The oscillator generates varying amplitude and phase signals, and the desired pose solver maps the amplitude and phase information into Cartesian space. The desired angle signals of the motors are calculated by inverse kinematics to control the position of the motors.

\begin{figure*}[t]%
	\centering
	\includegraphics[width=0.9\textwidth]{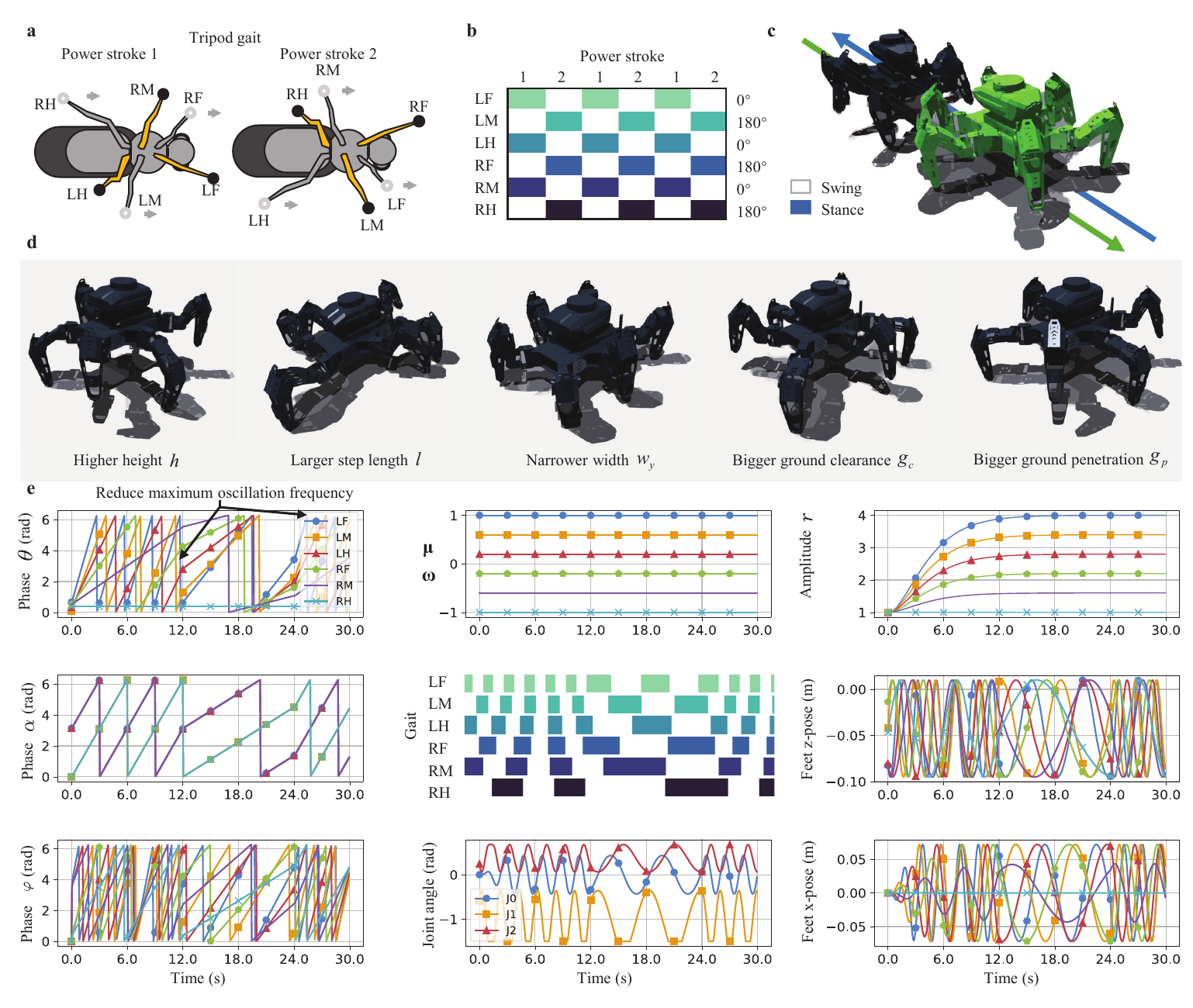}
	\caption{\textbf{Motion generation effect of the CPG module.} 
		\textbf{a}, the classic tripod gait of insects. There are two power strokes in a cycle. Each time, three legs stand on the ground (standing phase, orange legs, black circle), and the other three suspend in the air (swing phase, gray legs, gray circle). The arrow indicates the direction of movement. 
		\textbf{b}, Gait diagram of ideal tripod gait, standing (dark) and swinging (white) stages of each leg, and stroke number. 
		\textbf{c}, The robot moves forward or backward under the action of the CPG module embedded in the tripod gait phase.
		\textbf{d}, The robot can present different shapes by adjusting the morphological parameters of the CPG module through the lateral loop. 
		\textbf{e}, Through the $\bm{\mu},\bm{\omega}$ feedback of the ventromedial loop, The rhythm signals of the RG layer can be adjusted, and the PF layer can convert them into the poses of the end of legs under Cartesian coordinates to generate smooth joint signals.
	}\label{fig2}
\end{figure*}

Based on the above hierarchical structure, we further analyze the autonomous behavioral learning mode of the central nervous system. Due to the role of the genetic system, CPGs in the spinal cord are encoded at birth\textsuperscript{\cite{bib29}}, and rhythmic movement can be generated without acquired learning. The cerebellum has the ability for motor balance and coordination control. When learning new actions, these abilities are repeatedly strengthened, eventually leading to unconscious motor skills that can be repeatedly called by cortical regions\textsuperscript{\cite{bib23}}. Based on the above characteristics, mammals can devote more energy to the decision-making and learning process of the cerebral cortex on tasks to quickly solve various tasks and adapt to emergencies and new environments autonomously.

According to the characteristics of CPGs, we design a CPG module to generate autonomous motion signals without learning. We introduce the pre-encoded independent tripod gait phase to generate a stable phase signal. Based on the skill learning mechanism of the cerebellum, we propose a skill pre-training method based on unsupervised reinforcement learning and the CPG module, which can make the mid-level controller learn a variety of coordinated motor skills and can be repeatedly called by the high-level controller. According to the decision-making mode of the cerebral cortex on motion, we propose a two-stage multi-task learning method. The high-level controller can acquire the guided multi-task decision-making ability through reinforcement learning. Then, through distillation learning, the high-level controller learns to independently infer environmental features and motion direction from visual information to complete various tasks. This hierarchical independent learning mode can enhance the autonomy of the control system (the absence of higher levels will not affect the operation of lower levels), reduce the dependence on sensor information (the CPG module can also generate motion signals from internal oscillation in the case of sensor failure), and generate autonomous motion to cope with a variety of situations (solve multi-tasks and adapt to new environments).

\subsection{Results of the CPG module's motion generation}\label{subsec12}
\begin{figure*}[t]%
	\centering
	\includegraphics[width=0.9\textwidth]{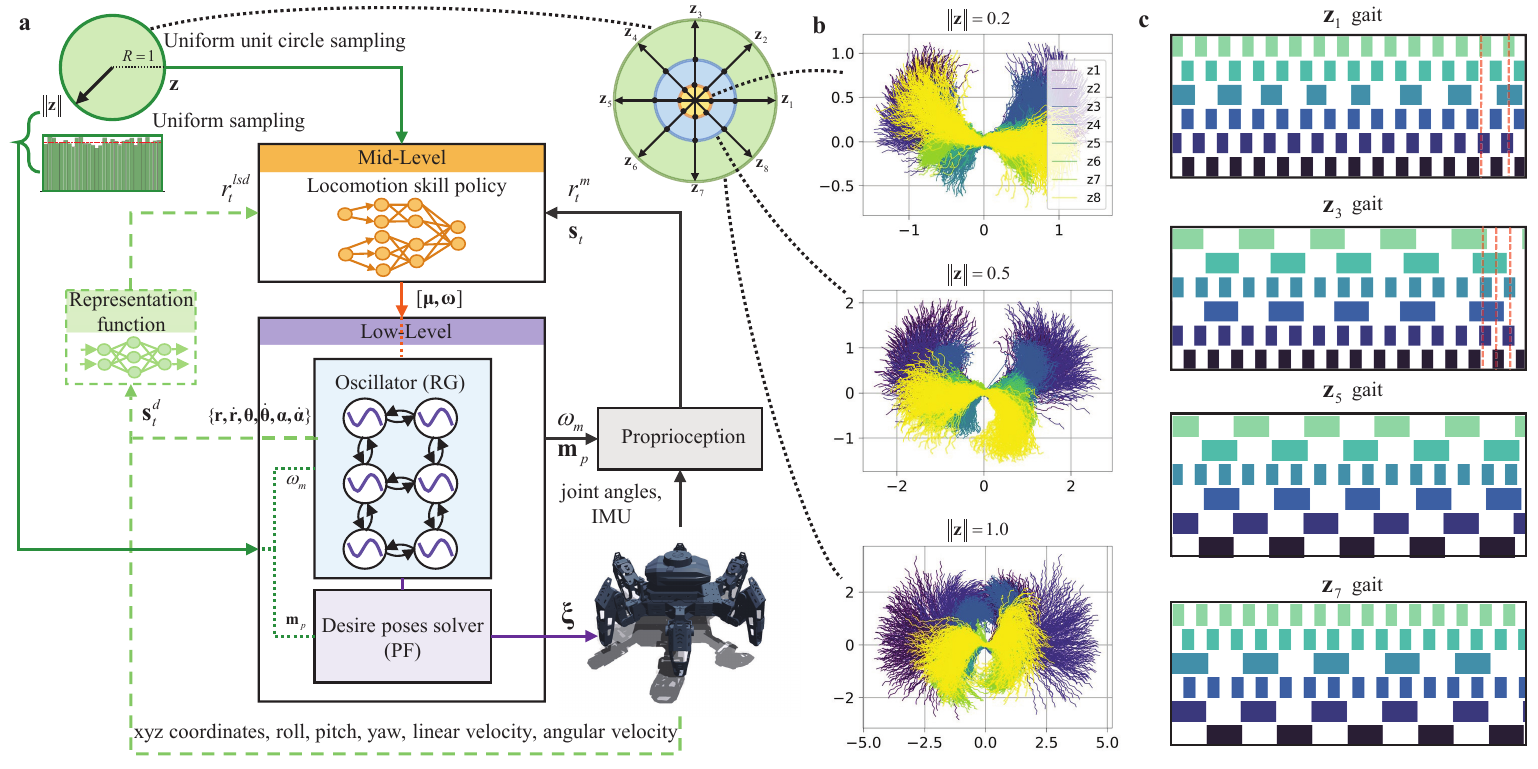}
	\caption{\textbf{Skill learning and regulation effects of the mid-level controller.} 
		\textbf{a}, Using unsupervised reinforcement learning to learn the mid-level controller.
		\textbf{b}, Under the action of different morphological parameters, different skill vectors make the robot generate various XoY plane motion trajectories.
		\textbf{c}, Under the action of one group of morphological parameters, the four skill vectors with a length of 1 form different gaits; dark color is the stance phase, and white is the swing phase.
	}\label{fig3}
\end{figure*}
The purple box in Fig. \ref{fig1}b shows the structure of the proposed CPG module. We use the amplitude-phase oscillator to simulate the RG layer and the desired pose solver to simulate the PF layer, and the high-level and mid-level controllers can regulate them directly and indirectly.

The oscillator consists of six elements acting on the robot's six legs respectively. We refer to the stable gait mode of insects\textsuperscript{\cite{bib30}} (Fig. \ref{fig2}a,b) and design the independent tripod gait phase $\bm{\alpha}$, which is only affected by the maximum oscillation frequency $\omega_m$, as the fixed component of the mixed phase $\bm{\varphi}$. The adjustable phase $\bm{\theta}$ provides the other component of $\bm{\varphi}$. This mixed phase enhances the autonomy of the CPG module and retains flexibility. Even if the higher control information is missing, the independent phase component can ensure the CPG module produces a stable forward or backward gait (Fig. \ref{fig2}c). In this mode, the oscillator can generate a regular amplitude signal $\bm{r}$, tripod gait phase $\bm{\alpha}$, and adjustable phase $\bm{\theta}$ (Fig. \ref{fig2}e). Amplitude $\bm{r}$ and adjustable phase $\bm{\theta}$ can be adjusted by $\bm{\mu},\bm{\omega}$ generated by the mid-level controller based on sensory feedback, thus making the CPG module generate more complex gait (the range of $\bm{\mu},\bm{\omega}$ in the figure is mapped to $[-1,1]$). Compared with the single-phase mode in refs \cite{bib20,bib21,bib22}, the introduction of the independent phase enables the oscillator to generate a stable phase signal when the higher signal has a boundary value (such as $\omega_i=-1$, light blue curve). This simple passive control module enhances the system's fault tolerance and further reduces its dependence on sensory information. In addition, the overall phase signals are adjusted by the maximum frequency $\omega_m$, thus controlling the overall speed of the robot. During 12-24s, the $\omega_m$ decreases, and the frequency of the three phases decreases simultaneously (Fig. \ref{fig2}e).

According to the amplitude and mixed-phase signals, the desired pose solver can calculate the desired Cartesian coordinates of the end of each leg and convert them into joint commands through inverse kinematics. Fig. \ref{fig2}e shows the Z and X coordinates curves and the smooth joint control signals. In addition, by changing the morphological parameters $\bm{m}_p=\{l,h,g_p,g_c,w_y\}$ (step length, height, support height, swing height, width) through the high-level controller, the robot can also show different shapes (Fig. \ref{fig2}d).

The results show that the CPG module is semi-autonomous and can generate stable motion mode by independent phase without external feedback. The incoming signals from higher-level controllers can affect the module, thus generating more complex control signals. These characteristics are the basis of subsequent multi-level active control.

\subsection{Results of the mid-level controller's skill learning and control}\label{subsec13}

The mid-level controller is a reinforcement learning policy composed of neural networks, responsible for the robot active skill regulation and gait control. It receives proprioception and skill vectors from the high-level controller. Then, it generates various differential parameters $\bm{\mu},\bm{\omega}$ for the oscillator, thus adjusting the CPG module to generate gait signals of different modes. External feedback $\|\bm{z}\|$ is proportional to the maximum internal frequency $\omega_m$, which can adjust the robot's speed. Combining the unsupervised reinforcement learning algorithm\textsuperscript{\cite{bib31}} and CPG module, we propose a skill pre-training method to learn the mid-level policy in simulation to find diversified motor skills (Fig. \ref{fig3}a). The CPG module can use low-dimensional parametric action space to generate high-dimensional joint control signals so that unsupervised learning can discover useful skills in robot systems with high degrees of freedom. (The learning process is reserved for the description of the Methods).

Fig. \ref{fig3}b shows the movement trajectories of the robot within 30 seconds of the XoY plane under the influence of different skills. To test the control effect of the mid-level policy, we select 24 skills with three different lengths in 8 different directions to drive 1024 robots with different morphological parameter combinations. It can be seen from Fig. \ref{fig3}b that the mid-level policy can generate movement trajectories in different directions under any combination of morphological parameters according to different skill vectors. With the increase of the skill length $\|\bm{z}\|$, the motion frequency of the robot legs also increases, resulting in a more distant motion mode.  Fig. \ref{fig3}c shows the gait maps corresponding to four skill vectors $\bm{z}_1,\bm{z}_3,\bm{z}_5,\bm{z}_7$ with a length of 1. It can be seen from the figure that the mid-level policy can produce various gait patterns under the effect of different skill vectors. For example, the robot will switch between biped and tripod gait under the action of $\bm{z}_1$. Under $\bm{z}_3$, the robot learn to switch between quadruped, tripod, and biped gait (Supplementary Section 6 further visualizes and analyzes the discovered skills). The results show that, without external data, the mid-level controller can effectively learn diversified skills and coordinate the limbs to produce multiple gait and trajectories like the cerebellum. The higher controller can later reuse these bounded continuous skills for rapid decision-making in various situations.

\subsection{Results of the high-level controller's multi-task learning and decision}\label{subsec14}

The high-level controller is also a reinforcement learning policy composed of neural networks, which is responsible for processing the robot's proprioception and environmental information, respectively, and then giving the appropriate high-level decision $\bm{a}_h=[\delta\bm{m}_p,\bm{z}] $ to actively command the lower levels to produce the appropriate movement pattern to deal with obstacles. $\delta\bm{m}_p$ directly changes the morphological parameters of the CPG module through the lateral loop. At the same time, $\bm{z}$ controls the mid-level controller to change the internal rhythm of the oscillator through the ventromedial loop. It is worth noting that $\|\bm{z}\|$ determines the maximum gait frequency $\omega_m$ of the robot, and this information is also transmitted directly from the lateral loop to the CPG module, which reflects the regulatory function of the basal ganglia on the willingness of the voluntary movement.

\begin{figure*}[t]%
	\centering
	\includegraphics[width=0.9\textwidth]{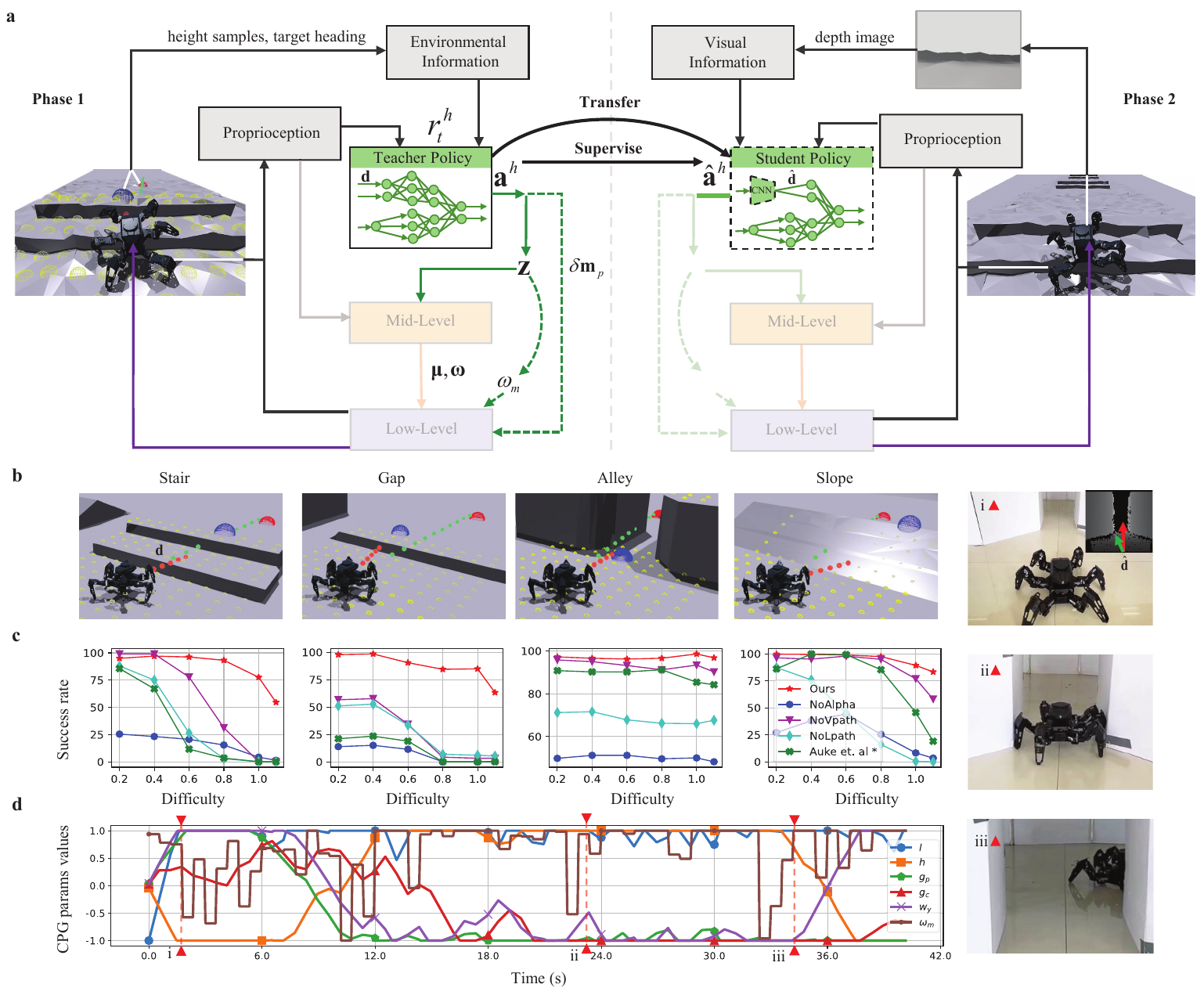}
	\caption{\textbf{Learning and decision results of the high-level controller.} 
		\textbf{a}, Multi-task reinforcement learning and distillation learning processes. 
		\textbf{b}, Schematic diagram of four simulation task environments. The robot acquires the height field information around the body through sensors. In an obstacle environment, target points will be randomly generated within a certain range to provide the robot with target heading directions.
		\textbf{c}, The success rates of different methods in different difficult tasks. Each baseline test uses 100 parallel robots to run for 60 seconds, and the average of 5 experiments is taken as the final success rate. The asterisk indicates recent work \cite{bib22}.
		\textbf{d}, Curve of morphological parameters (normalized result) under the action of the high-level policy while passing through the alley.
	}\label{fig4}
\end{figure*}

We propose a two-stage multi-task learning method, which allows the mid-level policy to remain unchanged after learning, and only learning the high-level policy quickly. In the first stage, we use the reinforcement learning algorithm to train multiple high-level policies in the multi-task target simulation environment ( Fig. \ref{fig4}b), allowing them to make decisions quickly. In the second stage, we use the distillation learning algorithm in the multi-task visual simulation environment to extract the learned high-level policies into a student policy, so that it can independently infer the environmental features and target heading directions according to the depth image, and then give appropriate high-level decision instructions (the two-stage learning process is reserved for the description of the Methods section). After completing the learning in the simulation environment, the student policy is directly deployed to the physical robot to perform four kinds of obstacle terrain crossing tasks autonomously.

We use different baselines for the first learning stage and conduct tests with increasing difficulty in the simulation (see Supplementary Section 2 for the configuration of different difficulties of each task) to evaluate the influence of different components of the proposed control framework on the control effect (including the introduction of the independent phase, hierarchical structure, and dual descending pathway).  Fig. \ref{fig4}c shows the success rate of each baseline. The settings of each baseline are as follows: 1) \textit{NoAlpha}: This removes the stable tripod gait phase in the CPG module, leaving the rest of the structure unchanged. 2) \textit{NoVpath}: This removes the mid-level policy and the ventromedial pathway, and the high-level policy directly provides all control signals reaching the CPG module through the lateral loop. 3) \textit{NoLpath}: This removes the lateral loop. The morphological parameters $\bm{m}_p$ cannot be adjusted.

Under the most difficult conditions of each terrain, the proposed method has a higher success rate (10\% -80\%). The effect of \textit{NoAlpha} is the worst. Due to the lack of a stable gait, the robot needs to learn to crawl first, so its learning efficiency is the lowest under the same number of iterations. The effect of \textit{NoVpath} is better than other baselines. Although it lacks the gait adjustment function of mid-level policy, it can directly adjust the morphological parameters according to the terrain to change the robot's shape to deal with obstacles. However, the robot must relearn the gait adjustment method when facing more difficult obstacles due to the lack of reusable motion skills. Therefore, with the increase the difficulty, the effect decreases rapidly. The effect of \textit{NoLpath} is inferior to \textit{NoVpath}. The robot cannot directly adjust the shape without the lateral loop. In the narrow alley task, the defect is very obvious. The robot cannot change its width, resulting in a low success rate. In addition, we compare with the state-of-art double-layer CPG-RL algorithm\cite{bib22}. Since no mid-level gait adjustment network exists, this method is similar to \textit{NoVpath}, and the success rate will drop sharply as the difficulty increases.

 Fig. \ref{fig4}d visualizes the changes of the five morphological parameters and the maximum frequency $\omega_m$ during the physical robot passing through the narrow alley. At 2th second, the robot finds the obstacle in front and starts to move. In 6-13 seconds, the high-level controller analyzes the characteristics of obstacles and makes the robot curl up and enter the alley. Its width $w_y$ decreases rapidly, its height $h$ increases, and its $\omega_m$ is high, making it move quickly in small steps. The squatting posture lasts until about the 34th second, and then the robot walks out of the alley, and the width returns to normal. (Supplementary Section 3 provides motion analysis of other tasks).

The above results show that the robot can use the learned skills to solve multiple tasks independently with the cooperation of high-level and lower-level controllers. The dual descending pathway structure enables the robot to have the ability of rapid decision-making and fine adjustment. The high-level controller can directly and quickly adjust the robot's shape (width, height, etc.) through the lateral loop according to the environmental information and provide skill instructions to the mid-level controller through the ventromedial loop to produce accurate gait and motion trajectory. The reusable generalization skills of the mid-level policy improve the learning speed of multiple tasks, while distillation learning enables the robot to have the ability of autonomous reasoning.

\subsection{Results of rapid recovery of movement in case of limb damage}\label{subsec15}

\begin{figure}[t]%
	\centering
	\includegraphics[width=0.45\textwidth]{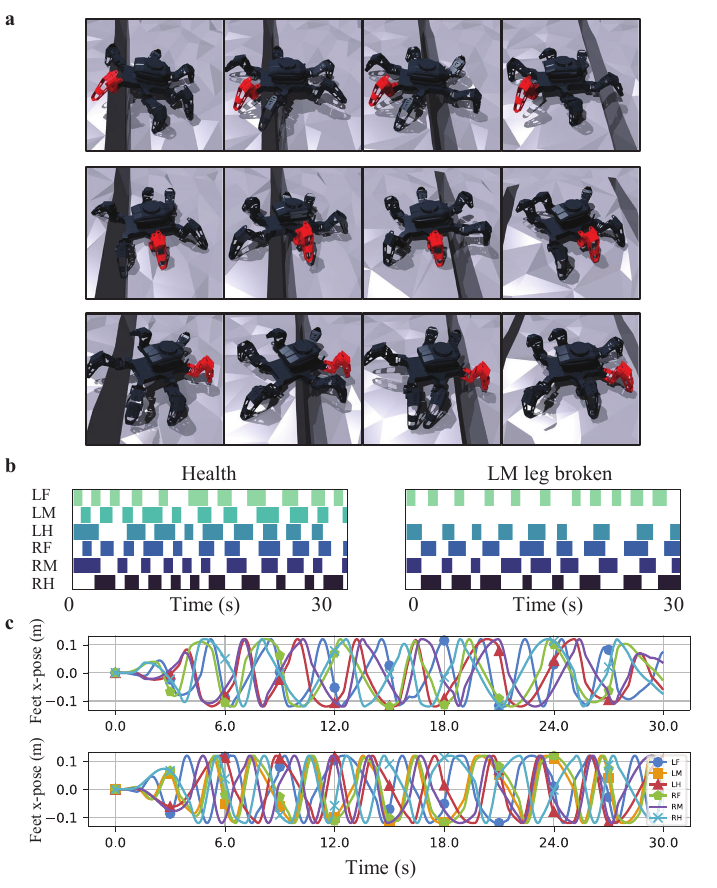}
	\caption{\textbf{The control effect of the robot in the case of a broken limb.} 
		\textbf{a}, When the front, middle, and hind legs of the robot are broken respectively, the robot can use the rest of the legs to cross the gap (the broken legs are marked in red, and we set the feedback and control signals of all joints of the leg to a fixed value to simulate the fracture). 
		\textbf{b}, Gait map of the healthy and left middle leg broken robot crossing the gap within 30 seconds.
		\textbf{c}, Changes of the X-coordinate at the end of each leg during process \textbf{b}.
	}\label{fig5}
\end{figure}
Animals exhibit rapid motor recovery after injury through various compensatory behaviors\textsuperscript{\cite{bib32}}. In the context of the bionic hierarchical structure, our novel method has the reliability of passive control systems, enables the robot to swiftly relearn its high-level controller in the event of limb damage, thereby coordinating different levels to restore its motion ability. The semi-autonomous CPG module further diminishes the control system's reliance on sensor information, enhancing its robustness and resilience.

To simulate the effect of limb damage, we randomly disable the control and feedback signals of a leg of the hexapod robot.  Fig. \ref{fig5}a demonstrates the robot's adaptability in crossing a gap (difficulty 0.6) when the robot's front, middle, and hind legs are broken, respectively. Regardless of which leg fails, the controller effectively guides the robot to adapt and utilize the remaining healthy legs to cross the gap and maintain a stable posture during crawling.  Fig. \ref{fig5}b further illustrates the gait changes of the robot in this task. After the failure of the middle leg, the other five legs of the robot adjust their gait and redistribute their workload to complete the obstacle crossing. Overall, the robot's gait frequency decreases, and the remaining legs have a longer time to swing and stance each time to maintain a stable crawling posture.  Fig. \ref{fig5}c presents the changes in X-coordinate at the ends of all legs of the robot during this process. In the healthy state, the trajectories of the robot's right front leg (RF), left middle leg (LM), right middle leg (RM), and left hind leg (LH) are very close, resulting in a coordinated pace. This coordination relationship is redistributed after the LM leg breaks down, leading the robot to adopt a new motion mode. Supplementary Section 4 provides a more detailed analysis of the impact of the different positions of the broken limb on recovery ability. 

\subsection{Results of unknown environment adaption}\label{subsec16}
\begin{figure*}[t]%
	\centering
	\includegraphics[width=0.9\textwidth]{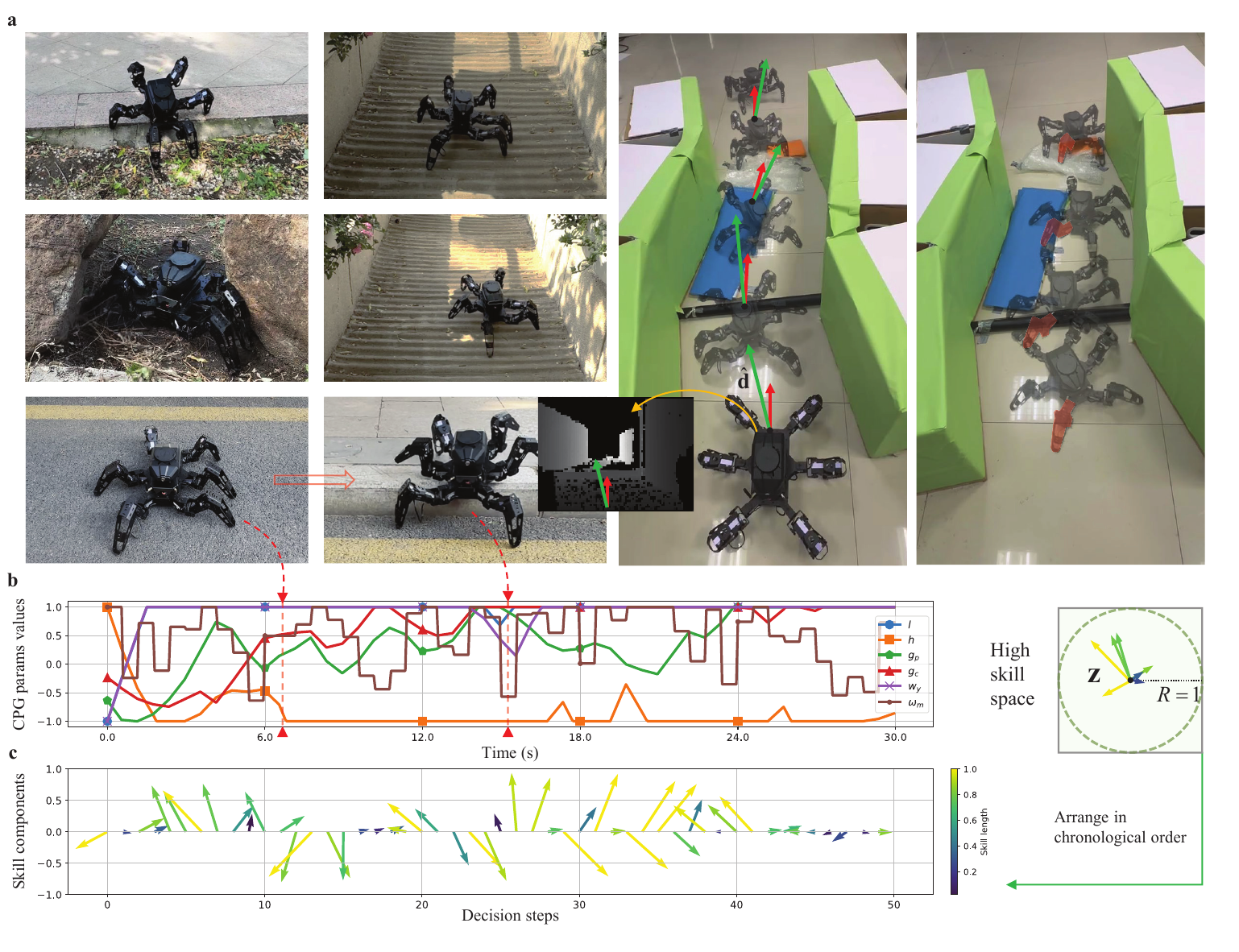}
	\caption{\textbf{Adaptation results of unknown environment.} 
		\textbf{a}, The robot exhibits strong mobility in new environments that have never been learned. It can move on the dirt ground, climb up or down a stone bridge, drill into irregularly shaped caves, climb the curb, and autonomously navigate in multi-obstacle combined terrain with a broken leg.
		\textbf{b}, The change of morphological parameters during climbing the curb. 
		\textbf{c}, The schematic diagram of skills changing with time given by the high-level controller of the \textbf{b} process. Each vector represents a skill $\bm{z}$, and the color of the vector represents its length.
	}\label{fig6}
\end{figure*}
After entering the unknown environment, animals can recombine the learned knowledge and solve new tasks instead of learning from scratch. The high-level active control system of the proposed method also has this advantage. As shown in  Fig. \ref{fig6}a, the hierarchical controller trained in the simulation environment is directly deployed to the robot system without modification, and its adaptability is tested in various unknown terrains (see Supplementary Video S5 for relevant videos). The robot can move independently on the outdoor dirt ground and cross the stone path to the opposite side through the high-level controller's autonomous inference and decision-making ability. It can climb the stone bridge steeper than the learned slope task. It can adjust its shape to drill into an irregularly shaped cave and quickly climb the curb. Moreover, it can navigate autonomously in multi-objective combined terrain without being affected by limb fracture (Supplementary Section 10 introduces the composition of the combined terrain and further analyzes the motion effect of the robot). 

 Fig. \ref{fig6}b shows the change of morphological parameters adjusted by the high-level controller over time while climbing the curb. After seeing the stone curb, the robot decided to reduce the height and increase the width to ensure stability during climbing. At the same time, increase $g_c$ so that the legs can be lifted higher to climb the stone platform. After climbing the stone platform, the maximum frequency $\omega_m$ decreases, maintaining low sports activity to save energy consumption. Fig. \ref{fig6}c shows the change of skill vector $\bm{z}$ generated by the high-level controller with the number of decision steps in this process. Under the effect of the hierarchical structure of the proposed method, the high-level controller will sample reusable skills from the skill space according to the environmental information at different times and combine them into continuous signals to guide the lower-level controllers to generate appropriate gait and trajectory to deal with the new environment.

\section{Discussion}\label{sec2}
We propose a general hierarchical learning control framework based on the structure and function of the central nervous system, which theoretically integrates active and passive control systems, and demonstrates in practice how to use hierarchical semi-active control systems to enhance the autonomous movement ability of robots. 
This framework provide new ideas and methods for the design and application of future robot control systems.
This framework refers to the structure and mechanisms of the mammalian central nervous system, including three interdependent hierarchical controllers, which can learn various motor skills and enable robots to use sensory feedback to produce autonomous motor behavior. Taking the complex hexapod robot system as the experimental object, we prove that this method can combine the reinforcement learning algorithm to quickly learn the controller in the simulation environment and directly deploy it to the robot system to handle complex tasks and independently solve new tasks. The low-level CPG module can independently generate stable motion mode and can be adjusted by the higher system. The mid-level controller can learn various skills to be repeatedly called by the high-level controller, while the high-level controller can quickly learn to handle various tasks. The dual descending pathway structure between the higher and lower controllers realizes fast and accurate motion control. 

This design concept based on the structure and mechanisms of the central nervous system significantly enhances the transparency and interpretability of the proposed control framework, giving it the advantages of the mammalian hierarchical nervous system\textsuperscript{\cite{bib1}}. Different sensory signals are specially processed at different levels. The high-level controller processes visual information and proprioception, while the mid-level controller only focuses on proprioception. This specialized processing method can promote the controller to fine-tune future tasks, reflecting the information factorization ability of the nervous system. The lower-level controllers can run semi-autonomously. Even without the high-level controller, the mid-level controller and the CPG module can produce many motion modes. The CPG module can even produce a stable gait without any external information. This passive control mode can reduce the interference between different levels, enhance the system's robustness, and reflect partial autonomy. The high-level controller can repeatedly call the mid-level and low-level controllers without relearning, which saves the learning cost and reflects the amortized control. The reward functions of the mid-level controller and the high-level controller adopt different settings. The mid-level controller adopts the intensive skill information reward, while the high-level controller uses the goal-conditioned global reward, reflecting the principle of modular objectives. The output of the high-level controller reaches the CPG module through two loops respectively, which is responsible for changing the overall shape and gait of the robot. Depending on the coupling relationship within the oscillator, the CPG module controls all limbs of the robot rather than independently controlling all joints. This coupling control can enhance the robot's motion coordination ability, reflecting the multi-joint coordination. The high-level controller generates control signals on a coarser time scale, improving decision-making efficiency. In comparison, the lower-level controllers generate accurate periodic signals on a finer time scale, reflecting the temporal abstraction. We believe that the research results are helpful to the development of more complex artificial intelligence systems. By studying the hierarchical control principle generated by the system, it is also helpful to understand the details of biological motion control.

In ref \cite{bib33}, a modular motion controller structure based on CPGs is proposed. It contains multiple passive control modules at different levels, each module can be learned separately, and then through different combinations, a hexapod robot can produce versatile behaviors to solve many tasks. However, the combination of modules still needs to be provided by the operator according to the actual situation, and this design logic weakens the autonomous movement ability of the robot. Ref \cite{bib6} proposed a three-layer generative model structure based on the principles of motion systems to simulate the deep temporal structure of human motion control, achieving multi-task active control of a humanoid robot in the simulation environment. However, this method relies on real-time feedback to generate control signals, and sensor or actuator failures can lead to unpredictable consequences. Our hierarchical learning control framework combines all the advantages of passive modular control structure and hierarchical active control structure. The passive motion generation mode of the CPG module based on spinal cord reduces sensory dependence, while the hierarchical active control structure of the reference central nervous system promotes the generation of autonomous behavior.

As mentioned above, the method proposed in this work improves the autonomous movement ability of robots and possesses excellent characteristics of the central nervous system. However, the absence of external force feedback from the foot or body may have adverse effects on the motion behavior when an irregular external force is applied to the robot. This disadvantage can be solved by introducing a local reflection loop to the CPG module\textsuperscript{\cite{bib35,bib36}}. By introducing a local conditioned reflex, the robot can automatically respond to external interference by intuition. On the other hand, although the CPG controller can produce smooth rhythm signals to enhance the stability of motion, it also makes it difficult for the robot to complete some explosive actions (such as jumping, galloping, etc.). One solution is introducing bias into the PF layer or directly outputting the joint position\textsuperscript{\cite{bib37, bib38}} to generate a burst control signal. Another solution is to use dynamic motion primitives (DMPs)\textsuperscript{\cite{bib39}} instead of the CPG module. DMPs have similar functions to CPGs and can generate discrete or rhythmic actions, often used for trajectory generation\textsuperscript{\cite{bib40,bib41}}. This discrete control signal is expected to make the robot produce more agile behavior with a sense of power.

The proposed control framework integrates the hierarchical structure and principles of the mammalian central neural system. However, the cognitive and reasoning ability of the high-level decision controller can be further improved. In future work, large visual or language models can be integrated into the decision-making level to upgrade the high-level controller \textsuperscript{\cite{bib42,bib43,bib44,bib45}}.
On the other hand, combining the mechanical model or structural features of robots to enable them to have reconstruction capabilities may lead to stronger robustness\textsuperscript{\cite{bib32,bibc}}. 
In addition, the agile movement of animals also comes from their precise body mechanics. Through the innovation of the mechanical structure of the robot, its movement ability can be further improved. For example, changing the position of the motor joints can make the mass ratio of the joints is closer to the biological structure. Using ``artificial muscle" instead of motors can make the mechanics of the robot more biologically compatible\textsuperscript{\cite{bib46,bib47}}.

This work generally shows how to develop a hierarchical learning control framework to help robots adapt to challenging environments by analyzing the mammalian central nervous system's hierarchical structure and mechanisms. It provides theoretical practice and engineering experience for designing an autonomous controller for a future robot system and further promotes the landing and application of a bionic autonomous controller.

\section{Methods}\label{sec3}
\small  
This section describes in detail the composition and learning methods of each part of the proposed hierarchical learning control framework and the application process for the hexapod robot (see Supplementary Section 1 for details of robot platform and physical simulation). The CPG module includes the oscillator and the desired pose solver. We introduce the differential equations of the oscillator and the internal stable phase embedding method, then show the desired pose solver's signal adjustment process and the robot motor's control method. Then, we give a detailed description of the pre-training method of the mid-level controller. Using the learned middle controller, we show how to get a high-level controller with autonomous decision-making ability through the two-stage learning process.

\subsection{Half-center rhythm generator layer}\label{sec31}
To generate the basic motion rhythm signal, we use the Hopf oscillation differential equations\textsuperscript{\cite{bib20,bib48}} to implement the RG layer of CPGs. The following first-order differential equations give the dynamic system:
\begin{equation}
	\begin{aligned}
		\dot{r}_i&=v_i  \\
		\dot{v}_i&=\frac{a^2}{4}(f({\mu_i})-r_i)-av_i  \\
		\dot{\theta}_i &= f({\omega_i})  \\
		\dot{\alpha}_i &= \frac{1}{2}{\omega_m}+\sum_j{r_jm_{ij}\sin(\alpha_j-\alpha_i-\Psi_{ij})} 
	\end{aligned}, \label{eq1}
\end{equation}
where $r_i$ is the amplitude of the oscillator, $v_i$ is the differential of amplitude, $\theta_i$ is the adjustable phase, $\alpha_i$ is the tripod gait phase, $a$ is a positive constant representing the convergence factor, $\mu_i$ and $\omega_i$ are amplitude and phase adjustment factors, $\omega_m$ is the maximum oscillation frequency. The oscillator finally produces a mixed phase $\varphi_i=(\alpha_i+\theta_i)$ ($i=1,2,...,6$ is the serial number of each leg). The coupling weight and bias between oscillation elements are $m_{ij}$ and $\Psi_{ij}$. They form an additive coupling term to generate an independent tripod gait for the robot, where $m_{ij}=1$, and the bias matrix $\bm{\Psi}$ is shown in the following formula
\begin{equation}
	\bm{\Psi}_{6*6}=2\pi\left[
	\begin{array}{cccccc}
		0 & 0.5 & 0 & 0.5 & 0 & 0.5 \\
		-0.5 & 0 & -0.5 & 0 & -0.5 & 0 \\
		0 & 0.5 & 0 & 0.5 & 0 & 0.5 \\
		-0.5 & 0 & -0.5 & 0 & -0.5 & 0 \\
		0 & 0.5 & 0 & 0.5 & 0 & 0.5 \\
		-0.5 & 0 & -0.5 & 0 & -0.5 & 0 \\
	\end{array}
	\right]        . \label{eq2}
\end{equation}

Due to the effect of the coupling term, the left front leg (LF), the left hind leg (LH) and the right middle leg (RM) of the robot are a group. Their $\alpha$ is the same, while the other three legs are another group, and their $\alpha$ lags $\pi$ rads. This setting makes the six legs form a tripod gait. On this basis, the mid-level controller can adjust the $\mu_i,\omega_i$ of each leg to directly change the amplitude $r_i$ and adjustable phase $\theta_i$ of the oscillator, then adjust the mixed phase $\varphi_i$ to make the CPG module produce different gaits. $f(\mu_i), f(\omega_i)$ are used to calculate the internal natural amplitude and frequency, where $f(\mu_i)=\mu_{min}+\frac{\mu_i+1}{2}(\mu_{max}-\mu_{min})$ and $f(\omega_i)=\omega_{min}+\frac{\omega_i+1}{2}(\omega_{max}-\omega_{min})$, they map  $\mu_i\in[-1,1]$,$\omega_i\in[-1,1]$ to $[\mu_{min}=1,\mu_{max}=4]$,$[\omega_{min}=0, \omega_{max}=\omega_{m}=u(\|\bm{z}\|)\Omega]$. $u$ is a linear mapping, which maps the $\|\bm{z}\|$ between 0 and 1 to $[0.2,1.0]$. $\Omega$ is a fixed value $8\pi$Hz, which can ensure that $\omega_m$ is always positive, thus ensuring that the independent tripod gait phase $\alpha$ is not affected by any external factors, and can always produce periodic tripod gait signals. This is different from previous work\textsuperscript{\cite{bib20, bib21, bib22}}. These methods add the external feedback signal $\omega_i$ and the coupling term directly and take them as the differential of a single phase. When the feedback signal is boundary value (such as $f(\omega_i)=0$), the only coupling term can not make the phase oscillate periodically, which makes the oscillator invalid.

We use the following formula to solve the state of the differential equations: 
\begin{equation}
	\begin{aligned}
		r_i^t=r_i^{t-1}+(\dot r_i^{t-1}+\dot r_i^{t})\frac{dt}{2}\\
		v_i^t=v_i^{t-1}+(\dot v_i^{t-1}+\dot v_i^{t})\frac{dt}{2}\\
		\theta_i^t=\theta_i^{t-1}+(\dot \theta_i^{t-1}+\dot \theta_i^{t})\frac{dt}{2}\\
		\alpha_i^t=\alpha_i^{t-1}+(\dot \alpha_i^{t-1}+\dot \alpha_i^{t})\frac{dt}{2}
	\end{aligned}, \label{eq3}
\end{equation}
where $dt=0.005$s.

\subsection{Pattern formation layer}\label{sec32}
To reshape the rhythm signal, we use the desired pose solver to realize the PF layer function of CPGs. After the oscillator generates $r_i,\varphi_i$, we calculate the desired pose of the end of each leg and then obtain the position under Cartesian Coordinates of the end of the leg, then convert it into the desired motor angles through the inverse kinematics, to generate the control signal of the motors. The end position of each leg is calculated as follows:
\begin{equation}
	\begin{aligned}
		& p_{x_i}=-{l}(r_i-1)\cos(\varphi_i)\\
		& p_{y_i}=L{w_y}\\
		& p_{z_i}=\begin{cases}
			-{h}+{g_c}\sin(\varphi_i),if \  \sin(\varphi_i)>0\\
			-h+{g_p}\sin(\varphi_i),otherwise\end{cases}
	\end{aligned} , \label{eq4}
\end{equation}
where $\varphi_i$ is the mixed phase, and $\{p_{x_i}, p_{y_i}, p_{z_i}\}$ is the position of the end of leg in the leg's local Cartesian Coordinates. $l$ is the step length, $L=l_1+l_2$, $l_1$, $l_2$, $l_3$ are the lengths of the three links of the robot coxa, femur and tibia, $w_y$ is the width adjustment variable, $h$ is the height of the robot, $g_c$ is the maximum ground clearance in the swing process, and $g_p$ is the maximum ground penetration in the support process. These parameters constitute the CPG morphological parameter set $\bm{m}_p=\{l,h,g_p,g_c,w_y\}$ (Fig\ref{fig1}.b shows the relationship between the above parameters and the gait). The high-level controller can directly adjust the shape of the robot by providing the deviation value $\delta \bm{m}_p=\{\delta l,\delta h,\delta g_p,\delta g_c,\delta w_y\}\in[-1,1]$. The adjustment process is as follows:
\begin{equation}
	\begin{aligned}
		l& \gets l+g(\delta l)\\
		h& \gets h+g(\delta h)\\
		g_p& \gets g_p+g(\delta g_p)\\
		g_c& \gets g_c+g(\delta g_c)\\
		w_y& \gets w_y+g(\delta w_y)\\
	\end{aligned} , \label{eq5}
\end{equation}
where $g(\delta x)=0.02(x_{max}-x_{min})\delta x$ maps the deviation to the range of specified parameters, which is given in Table.\ref{tab1}. When $l<0$, the foot trajectory rotates clockwise, and the robot moves backward. On the contrary, when $l>0$, the robot moves forward.
\begin{table}[h]
	\centering
	\caption{Range of morphological parameters}\label{tab1}%
	\setlength{\tabcolsep}{0.5mm}
	\begin{tabular}{@{}lllll@{}}
		\toprule
		 $l$  & $h$ & $g_p$ & $g_c$  & $w_y$\\
		\midrule
		 $[-0.12,0.12]$   & $[0.02,0.08]$  	 & $[0.03,0.06]$   & $[0.04,0.08]$  & $[1.1,1.5]$ \\
		\bottomrule
	\end{tabular}
\end{table}

After obtaining the position of leg end, we calculate the desired angles of $J_0$, $J_1$ and $J_2$ joints of each leg through the inverse kinematics model. (See Supplementary Section 11 for the calculation process). Through the PID controller inside the robot, the motors can be controlled to run to the specified angles.

\subsection{Skill learning of the mid-level controller}\label{sec33}
The mid-level reinforcement learning control policy can combine with the CPG module to form many coordinated motor skills. To achieve this, we use the parameterized neural network $\pi$ as the mid-level policy, and output $\bm{a}=[\bm{\mu}, \bm{\omega}]\in \mathbb{R}^{12}$ to adjust the internal amplitude and frequency of the oscillation, i.e. $\bm{a}_t\sim \pi(\bm{a}_t|\bm{s}_t,\bm{z})$, with a control frequency of 16.67Hz, according to the higher skill vector $\bm{z}$ and the robot's proprioception $\bm{s}_t$ (including 18 joint angles of the legs, rotational quaternions, angular velocities and linear accelerations information measured by the internal measurement unit (IMU), as well as the morphological parameters and maximum oscillation frequency of the CPG module).

Combining the unsupervised reinforcement learning method of work \cite{bib31} and the CPG module, we propose a new pre-training method, which can enable the mid-level policy to explore motor skills with excellent dynamic performance in the high degree of freedom system. The training process is shown in  Fig. \ref{fig3}a. The mid-level policy, CPG module, and robot environment form a closed loop. The CPG module provides the control signal, while the robot environment feeds back the state $\bm{s}_t$ and motion reward $r_t^m$ to the mid-level policy. During the learning process, the skill vector $\bm{z}$ is randomly selected from the unit circle using polar coordinates, while the morphological parameters are uniformly sampled within the range in Table.\ref{tab1}. By designing the parameterized state representation function $\phi$, we can map the homologous state $\bm{s}^d_t$ of the robot to the potential space and align it with the skills, then feed back the skill rewards $r^{lsd}_t$ to the mid-level policy. By maximizing the sum of rewards through the reinforcement learning algorithm SAC\textsuperscript{\cite{bib49}}, we can learn $\pi$ that applies to different morphological parameter sets. The algorithm maximizes the difference between the initial state and the final state under the condition of a certain skill, so that the policy can produce different trajectories under the effect of different skills, and its optimization objective is as follows:
\begin{equation}
	\begin{aligned}
		J_1(\pi) &=\mathbb{E}_{\bm{z},\tau}[(\phi(\bm{s}^d_T)-\phi(\bm{s}^d_0))^T\bm{z}] \\
		&=\mathbb{E}_{\bm{z},\tau}[\sum_{t=0}^{T-1}(\phi(\bm{s}^d_{t+1})-\phi(\bm{s}^d_{t}))^T\bm{z}] \\
		s.t.\quad &\forall x,y\in \mathcal{S}^d \quad \|\phi(x)-\phi(y)\|\le\|x-y\|\\
	\end{aligned} , \label{eq6}
\end{equation}
where $\bm{z}$ is the skill vector, $\tau$ is the trajectory generated by the policy under the effect of the skill, $\bm{s}^d_t$ and $\bm{s}_t$ are generated at the same time. However, different from $\bm{s}_t$, $\bm{s}^d_t$ mainly contains the robot's XYZ coordinates, IMU information (roll, pitch, yaw angles, linear velocities, and angular velocities) and oscillator's internal state $[\bm{r},\dot{\bm{r}},\bm{\theta},\dot{\bm{\theta}},\bm{\alpha},\dot{\bm{\alpha}}]$. The state representation function $\phi$ is also a learnable neural network responsible for mapping these information into the skill space and aligning skills with it. This asymmetric state structure makes the policy $\pi$ respond to the skill vectors associated with different oscillator's internal states while considering the proprioception. To avoid infinite $\phi(\bm{s}^d_t)$, 1-Lipschitz constraint is used on $\phi$. On the other hand, during the robot's movement, it is necessary to maintain the body's stability, so it is necessary to reduce the vertical speed of the robot. Therefore, another optimization objective is as follows:
\begin{equation}
	J_2(\pi) =\mathbb{E}_{\bm{z},\tau}[\sum_{t=0}^{T-1}-w_zv_z^2] , \label{eq7}
\end{equation}
where $w_z=0.1$ is the speed penalty weight, and $v_z$ is the speed of the robot in the z-axis direction. Thus, the reward per-step transition ($\bm{s}_t, \bm{a}_t, r_t, \bm{s}_{t+1}$) can be written as follows:
\begin{equation}
	r_t=r^{lsd}_t+r^{m}_t=(\phi(\bm{s}^d_{t+1})-\phi(\bm{s}^d_{t}))^T\bm{z}+(-w_zv_z^2) . \label{eq8}
\end{equation}

Considering the discount factor $\gamma$, the overall optimization objective of the mid-level policy is as follows:
\begin{equation}
	J(\pi) =\mathbb{E}_{\bm{z},\tau}[\sum^{T-1}_{t=0}\gamma^tr_t] , \label{eq9}
\end{equation}
we use SAC to optimize $\pi$ and use Stochastic Gradient Descent (SGD) to optimize $\phi$ (see Supplementary Section 5 for details of the algorithm and pseudo code).

Instead of sampling skills from Gaussian or von Mises distribution\textsuperscript{\cite{bib31, bib50}}, we uniformly sampled the skill vector $\bm{z}$ from the unit circle. Skills of different lengths can cover the whole space evenly. The mid-level policy can learn rich motion patterns under different oscillation frequencies. Another advantage is that the skill space can be conveniently used as the abstract action space of the high-level policy. In this regard, we design the following random skill generator:
\begin{equation}
	\begin{aligned}
		&R\sim U(0,1)\\
		&\beta\sim U(0,2\pi)\\
		&z_x =\sqrt R \cos(\beta)\\
		&z_y =\sqrt R \sin(\beta) 
	\end{aligned} \label{eq10}
\end{equation}
Through polar coordinate transformation, we can get the vector $\bm{z}=(z_x,z_y)$ uniformly distributed in the unit circle (See Supplementary Section 7 for the proof of uniform sampling).

\subsection{Multi-task reinforcement learning of the high-level controller}\label{sec34}
To make the high-level reinforcement learning control policy have the ability of autonomous decision-making, we propose a two-stage multi-task reinforcement learning method. The first stage is carried out in the multi-task simulation environment ( Fig. \ref{fig4}b). We use the parameterized neural network $\eta$ as the high-level policy. It accepts the robot's proprioception $\bm{s}_t$ and the environmental information $\bm{s}^e_t$ from the environment (including the height sampling points centered on the robot and the heading directions from the robot to the two target points). It generates the high-level decision action $\bm{a}^h=[\delta \bm{m}_p,\bm{z}]\in \mathbb{R}^7$, i.e. $\bm{a}^h_t \sim \eta(\bm{a}^h_t|\bm{s}_t,\bm{s}^e_t)$. Use the learned skills to control the robot movement, we can get the environmental reward $r^h_t$. Due to the time abstraction of the hierarchical structure, the action execution frequency (1.67Hz)  of the high-level policy is only $\frac{1}{10}$ of that of the middle level, which saves computational resources and improves efficiency.

The learning process is shown in  Fig. \ref{fig4}a phase 1. After learning the mid-level policy, fix it and only learn the high-level policy. Each time the high-level policy interacts with the environment, it will receive a reward $r^h_t$. Through the model-free reinforcement learning algorithm (SAC), the total reward of each task can be maximized, and the learning process can be completed in a short time (about $\frac{1}{3}$ of the learning time of the mid-level policy). The overall optimization objective of $\eta$ can be written as:
\begin{equation}
	J(\eta) =\mathbb{E}_{\tau}[\sum^{T-1}_{t=0}\gamma^tr^h_t], \label{eq11}
\end{equation}
where $\tau$ is the trajectory generated in each episode. Due to the time abstraction, the number of transitions collected during high-level policy training will be reduced. To solve the problem of low sample efficiency, we use step-conditioned critical SAC structure\textsuperscript{\cite{bib51}}, (see Supplementary Section 8 for details of the algorithm and pseudo code). The reward function uses the following form:
\begin{equation}
	r^h=d_l\times(w_v\times r_v+w_d\times r_d+w_b\times r_b+w_s\times r_s+w_T\times r_T) , \label{eq12}
\end{equation}
where $d_l$ is the difficulty level ($d_l = 1,2,3,4,5$). The higher the difficulty level, the richer the reward. To enable the robot to adapt to obstacles quickly, we adopt the dynamic learning mode of course learning and divide the obstacle environment into different difficulty levels. When the robot reaches the final goal, it will be transferred to a more difficult environment. If the robot cannot achieve half of the goal, it will be transferred to a simpler environment. The total reward mainly includes five sub rewards: speed tracking reward $r_v$, direction tracking reward $r_d$, balance reward $r_b$, collision reward $r_s$ and completion time reward $r_T$. The weight set of sub rewards is $\{w_v,w_d,w_b,w_s,w_T\}$.

To enable the robot to climb in any direction and overcome obstacles instead of moving around obstacles, we use the target points in the world coordinates to calculate the desired heading directions, which can be written as 
\begin{equation}
	\boldsymbol d=\frac{\boldsymbol {g} -\boldsymbol{x}}{\| \boldsymbol {g} -\boldsymbol{x}\|}, \label{eq13}
\end{equation}
where $\bm{g}$ is the coordinates of the target points, and $\bm{x}$ is the coordinate of the robot in the world coordinates. The following speed tracking reward $r_v$ encourages the robot to move towards the target points
\begin{equation}
	r_v = \min(<\boldsymbol d,\boldsymbol v>,v_{max}), \label{eq14}
\end{equation}
where $\bm{v}$ is the speed of the robot in the world coordinates, $v_m$ is the maximum running speed of the robot, here is 0.3m/s.

Use the following direction tracking reward $r_d$ to prompt the robot to quickly adjust its motion direction
\begin{equation}
	r_d=\exp(-|\boldsymbol d-\boldsymbol d_c|), \label{eq15}
\end{equation}
where $\bm{d}_c$ is the current heading directions vector of the robot in the world coordinates.

Use the following balance reward $r_b$ to punish the robot for its vertical movement and avoid the robot shaking up and down during the movement
\begin{equation}
	r_b=- v_z^2, \label{eq16}
\end{equation}
where $v_z$ is the speed of the robot in the vertical direction.

Use the following collision reward $r_s$ to reduce the number of collisions between the foot and the obstacle during robot movement. 
\begin{equation}
	{r_s=}
	\begin{cases}-1 ,if \ |f_{xy}|>4|f_z|\\
		0 , otherwise
	\end{cases},  \label{eq17}
\end{equation}
where $f_{xy}$ is the resultant force in the XY direction and $f_{z}$ is the force in the Z direction.

Use the following time reward $r_T$ to urge the robot to complete the task quickly
\begin{equation}
	r_T=-0.1. \label{eq18}
\end{equation}

The weight set differs for the four obstacle environments to improve the learning effect. For stair climbing, we focus on the robot's ability to climb quickly, so the weight of the balance reward is reduced, and its reward weight set is $\{w_v=1.0,w_d=1.0,w_b=0.5,w_s=1.0,w_T=1.0\}$. For crossing the gap, we focus more on its balance ability, and the reward weight set is $\{w_v=1.0,w_d=1.0,w_b=2.0,w_s=1.0,w_T=1.0\}$. For crossing the narrow alley, the robot should move quickly and avoid hitting the walls on both sides, so the weight of speed and collision reward has been improved, $\{w_v=2.0,w_d=1.0,w_b=2.0,w_s=2.0,w_T=1.0\}$. When the robot climbs over the slope, it needs to adjust its heading directions quickly on the slope to avoid falling, so its direction tracking reward has a higher weight, $\{w_v=1.0,w_d=1.5,w_b=1.0,w_s=1.0,w_T=1.0\}$.

\subsection{Distillation learning of the high-level controller}\label{sec35}
In the first stage of the learning process, the robot uses the surrounding height field and target heading directions as the external environmental information in the simulation environment, which can accelerate the learning and sampling process. However, the real robot must perceive the external environment through the camera. Therefore, in the second stage, distillation learning is used in the multi-task visual simulation environment, and multiple high-level policies learned are extracted into a student policy so that it can independently infer the environmental features and target heading directions according to the depth image, and then give appropriate high-level decision instructions. We use the distillation learning method of teacher-student to collect virtual image information for training in the simulation.

The distillation learning process is shown in  Fig. \ref{fig4}a phase 2. We use multiple high-level neural networks for different tasks as the teacher policies and a neural network that can receive image input as the student policy to learn in the most difficult obstacle environment (see Supplementary Section 9 for the structure of the student network). The student policy needs to predict the height information near the robot and the target heading directions vector $\hat{\bm{d}}$ from the depth image and generate the predicted action $\hat{\bm{a}}^h_t$ through the predicted information. If the heading directions vector predicted by the student policy is inaccurate initially, the direct use of the predicted vector as the state may lead to catastrophic distribution drift. To deal with this problem, we refer to ref \cite{bib38} and use Dagger\textsuperscript{\cite{bib52}} to train the student policy. Specifically, we use the mixed heading directions of the teacher and student as the target heading directions. The heading directions are calculated as follows:
\begin{equation}
	{\boldsymbol s^e_{\theta} =}
	\begin{cases}
		\hat{\boldsymbol d} ,if \ |\boldsymbol d-\hat {\boldsymbol d}|<0.6\\
		\boldsymbol d , otherwise 
	\end{cases} \label{eq19}
\end{equation}
where $\hat{\bm{d}}$ is the heading directions vector predicted by the student policy, and $\bm{d}$ is the actual calculated heading directions vector. The predicted heading directions are adopted when the difference between the two is within the allowable range. Otherwise, the actual heading directions are adopted. $\bm{s}^e_{\theta}$ is the heading directions vector of observations obtained by the student policy.

During distillation, supervised learning is used to train the student policy. The loss function is as follows:
\begin{equation}
	loss=\frac{1}{N}\sum_{i=1}^N(\|\hat{\boldsymbol d}-\boldsymbol d\|+\|\hat{\boldsymbol a}^h -\boldsymbol a^h \|), \label{eq20}
\end{equation}
where $\hat{\bm{a}}^h$ is the high-level action predicted by the student policy, and $\bm{a}^h$ is the guiding action generated by teacher policies. In the training process, we add random noise to the depth image obtained in the simulation environment to enhance the robustness of the policy to image input in the physical environment.

\section{Acknowledgements}\label{sec4}
Parts of the  Fig. \ref{fig1}a by using pictures from Servier Medical Art. Servier Medical Art by Servier is licensed under a Creative Commons Attribution 3.0 Unported License (https://creativecommons.org/licenses/by/3.0/).

\normalsize  

\bibliography{bib52}
\end{document}